\documentclass[lettersize,journal]{IEEEtran}
\usepackage{amsmath,amsfonts}
\usepackage{algorithmic}
\usepackage{algorithm}
\usepackage{array}
\usepackage[caption=false,font=normalsize,labelfont=sf,textfont=sf]{subfig}
\usepackage{textcomp}
\usepackage{stfloats}
\usepackage{hyperref}

\usepackage{verbatim}
\usepackage{graphicx}
\usepackage{cite}
\usepackage{authblk}
\begin{document}

\title{Comprehensive Study on Performance Evaluation and Optimization of Model Compression: Bridging Traditional Deep Learning and Large Language Models}

\author[]{Aayush Saxena$^\dag $}
\author[]{Arit Kumar Bishwas$^{\dag^*}$}
\author[]{ Ayush Ashok  Mishra}
\author[]{Ryan Armstrong}

\affil[]{Innovation Hub, PricewaterhouseCoopers
 USA}



\maketitle
\def\thefootnote{\dag}\footnotetext{Equal Contributors.}
\def\thefootnote{*}\footnotetext{ Corresponding Author .}
\begin{abstract}
Deep learning models have achieved tremendous success in most of the industries in recent years. The evolution of these models has also led to an increase in the model size and energy requirement, making it difficult to deploy in production on low compute devices. An increase in the number of connected devices around the world warrants compressed models that can be easily deployed at the local devices with low compute capacity and power accessibility. A wide range of solutions have been proposed by different researchers to reduce the size and complexity of such models, prominent among them are, Weight Quantization, Parameter Pruning, Network Pruning, low-rank representation, weights sharing, neural architecture search, knowledge distillation etc. In this research work, we investigate the performance impacts on various trained deep learning models, compressed using quantization and pruning techniques. We implemented both, quantization and pruning, compression techniques on popular deep learning models used in the image classification, object detection, language models and generative models-based problem statements. We also explored performance  of various large language models (LLMs) after quantization and low rank adaptation. We used the standard evaluation metrics (model’s size, accuracy, and inference time) for all the related problem statements and concluded this paper by discussing the challenges and future work.\end{abstract}



\section{Introduction}
\IEEEPARstart{D}{eep} learning based solutions have achieved great success to the problems in almost every domain. This success has been achieved at the cost of high memory and compute requirements which makes deployment of Deep Neural Network based models on edge devices a difficult task. There has been a drastic increase in the model size and data that is being used to build these SOTA solutions. For example, the natural language processing models have gone from a few billion parameters trained on tens of billions of
token data to hundreds of billions or trillions of parameters trained on trillions of token data \cite{1}.
The commonly used models for computer vision tasks also have around a few billion FLOPS along with several million parameters \cite{2}.
Training a neural network on a specific dataset requires capturing as much information as it can to perform well on the given task. For this purpose, the weights and activations of a neural network are initialized using single-precision floating point format i.e., float32 values which require 32 bits of memory to store the value as defined by the IEEE 754-2008 standard \cite{3}. While inference, these parameters are loaded into the local memory and operations are performed on float32 data by the processor. During this time, it may not be essential to capture all the information precisely to perform the given task and lower precision value might also give similar results. Hence, these model parameters can be quantized to different values to reduce the size of the model and decrease the inference time. Quantization is the process of mapping values from a large set to a smaller set of values while having minimum loss of information. In deep learning, models can be quantized from float32 values to int8 values by using different quantization algorithms. Deep learning frameworks like TensorFlow \cite{4}, PyTorch \cite{5} and Caffe \cite{6} provide a framework to optimize models using quantization techniques. In all the frameworks, quantization methods can broadly be divided into two: Quantization Aware Training and Post-Training Quantization. In post training quantization using TensorFlow, the models can be optimized for latency and size using Dynamic Range Quantization, Float16 Quantization and Int8 optimization by converting it to tf-lite architecture \cite{4}. TF-lite is Google’s open-source machine learning framework that converts a pretrained model in TensorFlow to a special format that can be optimized for speed and storage and can directly be deployed on devices such as mobiles (android and iOS), desktop and edge devices. However, the quantized version of a model might not always result in a desirable output and may considerably lower the performance as compared to the original.
Also, the billions of parameters in a pre-trained model do not contribute equally to the performance of the model and some of the neurons or layers can be eliminated to increase the sparsity while least affecting the original performance. This is called pruning of the Neural Network, where we eliminate the least contributing layers or neurons or channels in a pre- trained model. There are many different types of pruning but prominent among them are structured pruning and unstructured pruning. The structured pruning involves selective removal of larger parts of the network such as a layer or a channel while the unstructured pruning can be understood as finding and removing the less salient connections in the model. Deep learning frameworks like Tensorflow and PyTorch provide pruning features to make the model smaller.

\section{Related Work}
A wide range of researchers are actively working on solving the problem of deploying DNN based models on devices with limited compute capacity. Reducing the size of the model in order to minimize the daunting computational cost by using integer-only arithmetic operations while inference on mobile devices has been explored by Jacob \textit{et al.} \cite{7}. Quantizing convolutional neural networks for inference with integer weights and activations has been presented by Krishnamoorthi \textit{et al.}\cite{8}, which demonstrates per-channel quantization of weights and per-layer quantization of activations to 8-bits of precision post-training produces classification accuracies within 2\% of floating-point networks for a wide variety of CNN architectures. For benchmarking the performance of these methods, Dubhir  \textit{et al.} \cite{9} have evaluated the various features of the quantization process supported in Pytorch and Tensorflow on CNN and GNN based Recommendation models. To compare and rank the AI capabilities of the devices, Luo \textit{et al.} \cite{10} proposed two unified metrics as AI score: Valid Images per Seconds (VIPS) and Valid FLOPs per second (VOPs). Apart from these, TFLite also provides a benchmarking tool that currently measures and computes statistics for initialization time, inference time of warm-up state, inference time of steady state, memory usage during initialization time and overall memory usage \cite{11}. However, these benchmark tools are available as benchmark apps for Android and iOS and as native command line libraries while sharing the same core performance measurement logic \cite{11}.
In their paper, "What is the state of the Neural Network Pruning?", Blalock \textit{et al.} \cite{12} aggregate the results of 81 research papers and conclude that the community suffers from a lack of standardized benchmarks and metrics. To address the issue of benefits of pruning across different architectures, they plotted the reported accuracies and compression/speedup levels of pruned models on ImageNet alongside the same metrics. It concludes that pruning can improve the time or space vs accuracy tradeoff of a given architecture, sometimes even increasing the accuracy. In Data-free Parameter Pruning for Deep Neural Network, Srinivas \textit{et al.} \cite{13} proposes removing one neuron at a time rather than removing individual weights. Using systematic removal of redundant neurons, they achieve upto 35\% of sparsity on the Alexnet without significantly affecting the performance.
\begin{figure}[!t]
\centering
\includegraphics[width=2.5in]{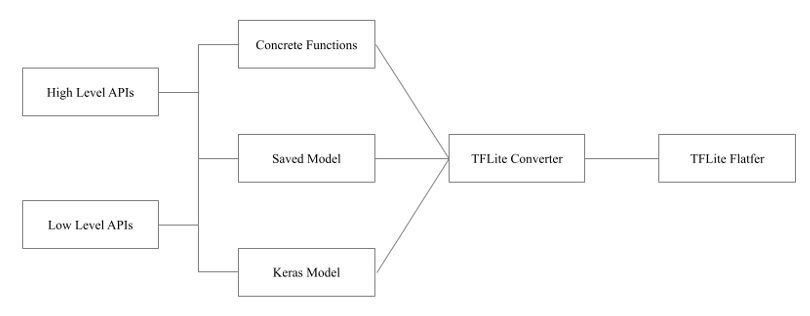}
\caption{Model format supported for TFLite converter}
\label{fig:img1}
\end{figure}

\section{Methods}
\subsection{Quantizing tensorflow models}

Post-training quantization is a conversion technique that can reduce model size while also improving CPU and hardware accelerator latency, with little degradation in model accuracy. The pre-trained TensorFlow model can be converted into TensorFlow Lite format using the TensorFlow Lite Converter which generates an optimized FlatBuffer \cite{14} format identified by .tflite extension \cite{4}. In addition to creating FlatBuffers, TFLite converter can also apply optimizations to the model which reduces the size of the model or inference time or both.
Quantization is one of the optimization methods that can be applied to the TFLite Flatbuffer. In the dynamic range quantization, the weights of the model are statically converted from a floating-point number to an integer value which reduces the models size by approximately x4 times. At inference, weights are converted from 8-bits of precision to floating point and computed using floating-point kernels while the activations are converted to integer values. This conversion is done once and cached to reduce memory. With this optimized model we can only use CPU for inference and GPU or TPU cannot be used.
In float16 quantization, the size of a model is reduced from float32 value to float16 as defined by the IEEE standard for 16-bit floating point number \cite{15}. As this uses the maximum number of bits for representing as compared to the other two quantization methods, it causes minimal loss to the accuracy. It supports delegates that can utilize GPUs and hence it performs faster as compared to the float32 model.
\begin{figure}[!t]
\centering 
  \includegraphics[width=3.5in]{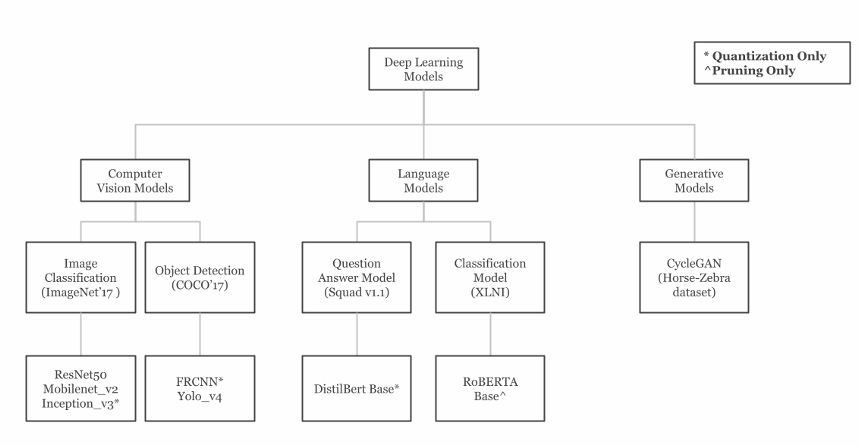}
  \caption{Quantization Method Comparison}
  \label{fig:img2}
\end{figure}

The int8 quantization converts all of the operations of the model to int8 value. It requires a set of sample data points to calibrate or estimate the range of all the floating-point tensors in the given model. In our experiments we have used 100 data points from the training data as the representative dataset while optimizing the model for int8 quantization. This type of optimization is preferred for deploying models on microcontrollers, Edge TPU or CPU, as they require integer operations to accelerate the performance. Table \ref{tab:table1} shows the benefits of using each of these quantization techniques along with hardware that supports them.

\begin{table}[!t]
\caption{Quantization Method Comparison \label{tab:table1}}
\centering

\scalebox{0.6}{\begin{tabular}{|c||c||c|} 
 
 \hline
    Techniques  & Benefit & Hardware \\  
 \hline
Dynamic Range Quantization  & 4x smaller, 2x-3x speed & CPU\\
 \hline 
Float16 Quantization & 2x smaller, GPU acceleration & CPU, GPU \\
 \hline 
 Full Integer Quantization & 4x smaller, 3x+ speed & CPU, Edge TPU, Microcontrollers \\
\hline
\end{tabular}}
\end{table}

\subsection{Pruning Deep Learning models}
Deep neural networks are dense and overparameterized which leads to higher memory footprint and power requiremnet to deploy those models. Pruning methods can be of different forms depending upon the desired output but are generally classified into structured and unstructured pruning \cite{16}. In unstructured pruning, parameters which are least important in the network are removed which makes the model sparse and this sparse model is again retrained, and this iterative process continues until we reach the desired level of pruning. TensorFlow \cite{17} provides magnitude based weight pruning that gradually reduces the insignificant weights of a model to zero and increases the sparsity. Sparse models can be compressed easily and zero weights can be eliminated during inference to reduce the latency of the model. The structured pruning works at filter/channel level in convolutional neural networks. The complete channels are removed as per the ranking of importance score calculated with respect to loss function. For pruning image classification and object detection models we used the tensorflow pruning api \cite{17} whereas for pruning language models i.e. XLM -Roberta and Bert-small we used strucutured pruning method proposed by Yang \textit{et al.} \cite{18}. For cyclegan compression we used pruning and knowledge transfer based technique proposed by Wang \textit{et al.} \cite{19}.

\begin{figure}[!t]
\centering 
  \includegraphics[width=2.5in]{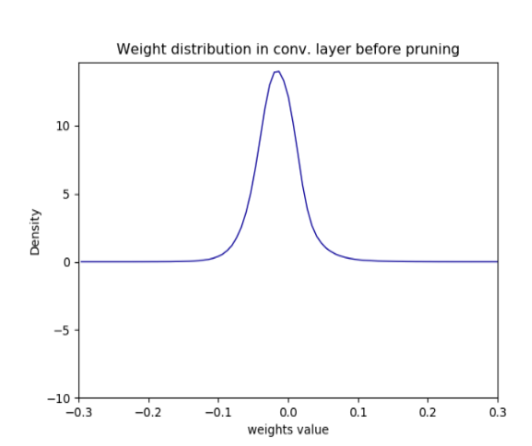}
  \caption{Weight Distribution in a Conv. Layer Before Pruning}
  \label{fig:img10}
\end{figure}

\begin{figure}[!t]
\centering 
  \includegraphics[width=2.5in]{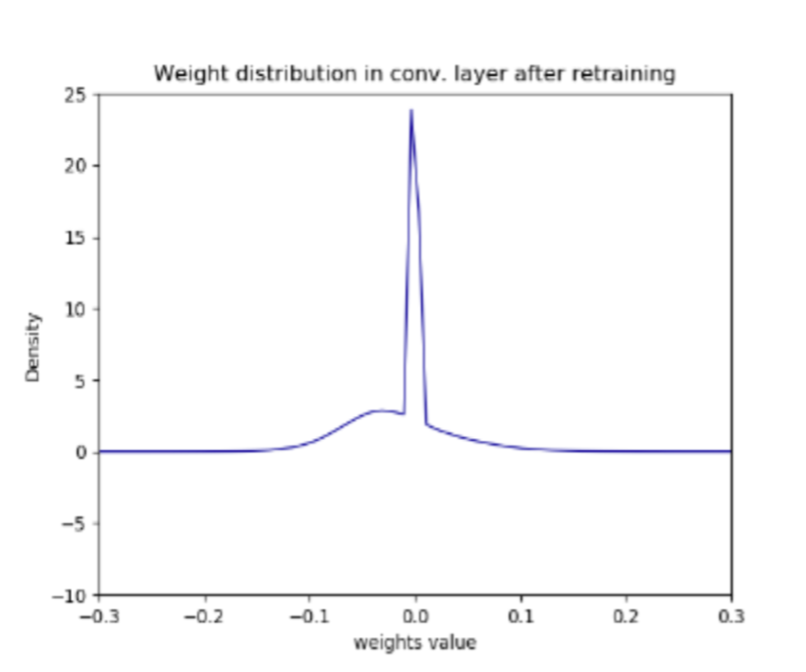}
  \caption{Weight Distribution in a Conv. Layer After Pruning}
  \label{fig:img11}
\end{figure}

\begin{figure}[!t]
\centering
 \includegraphics[width=2.5in]{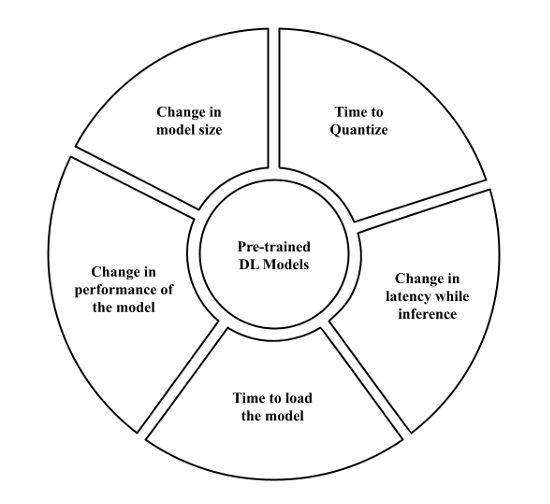}
  \caption{Metrics Monitored During Experiments}
  \label{fig:img3}
\end{figure}

\begin{figure}[!t]
  \centering
  \includegraphics[width=2.5in]{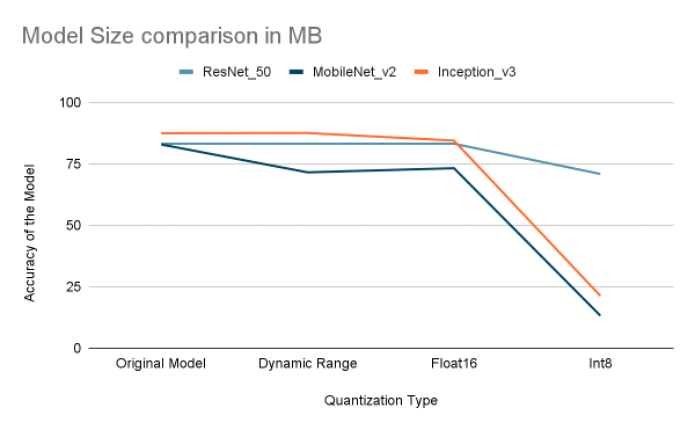}
  \caption{Image Classification - Model Accuracy Comparison in Percentage}
  \label{fig:img4}
\end{figure}

\subsection{Choosing the models and dataset to test quantization and pruning}
To test the effects of pruning or post training quantization on model’s performance we chose seven different pretrained models with four different problem statements. For the image classification task, we used pretrained ResNet50, MobileNet\_v2 and Inception\_v3(quantization only) models, all pre- trained on ImageNet \cite{20} dataset and taken from keras library \cite{21}. While the motivation behind creating the Inception network was to reduce the computational cost, ResNet focused more on computational accuracy and MobileNet was designed to solve computer vision problems on the computationally limited platforms. These factors made them ideal for comparing the effects of post training quantization on pretrained models.

For object detection tasks, we tested YOLO\_v4 \cite{22} and FRCNN \cite{23} (for quantization only) with ResNet50 backed models, both pre-trained on COCO dataset \cite{24}. Although both these models use an anchor box-based network structure, YOLO differs from FRCNN in that it makes classification and bounding box regression at the same time. Faster RCNN on the other hand detects small objects well since it has nine anchors in a single grid, however it fails to do real-time detection with its two-step architecture \cite{25}.
For quantizing language models, we used DistilBert Base Uncased \cite{26} which is a smaller, faster and distilled version of BERT and has half the number of parameters than the original. The pre-trained model on the Squad v1.1 dataset \cite{27} was taken from the hugging face library \cite{26} and was tested on validation data. In the generative models we took an open source CycleGAN \cite{28} model trained on the horse to zebra dataset \cite{29}. For pruning language models, we used XLM-RoBERTA \cite{31} trained on 2.5Tb of filtered CommonCrawl data containing 100 languages, BERT-small\cite{51} trained on disaster tweets dataset\cite{50} and GPT-2 small\cite{49} finetuned on E2E NLG \cite{novikova2017e2e}.

\begin{figure}[!t]
\centering 
  \includegraphics[width=2.5in]{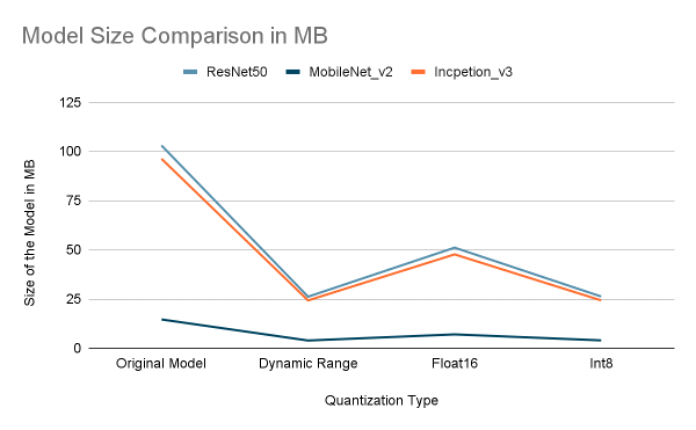}
  \caption{ Image Classification - Model Size Comparison in MB}
  \label{fig:img5}
\end{figure}
\subsection{Metrics Monitored}
All of the seven models were quantized and tested on MacBook Pro 2019 powered by 2.3 GHz 8 core Intel i9-9880H \cite{30} processor and 16GB 2400MHz DDR4 memory. To test the deployability of a model the important parameters to know are its inference time, performance and the size. Hence, we monitored the changes in these parameters for different quantized models. Along with this we also observed the time it takes to optimize the models using different quantization methods and load these models to local memory for computations. For model pruning, we retrained the models on Azure NC6s V3 \cite{48} instance with 6 vCPUs, RAM of 112GiB and a single GPU of 16GiB \cite{48}. For compression of GPT-2 small we used Azure NC12s v2 instance with 12 vCPUs, RAM of 224 GiB and two GPUS of 16GiB.

\begin{figure}[!t]
\centering 
  \includegraphics[width=2.5in]{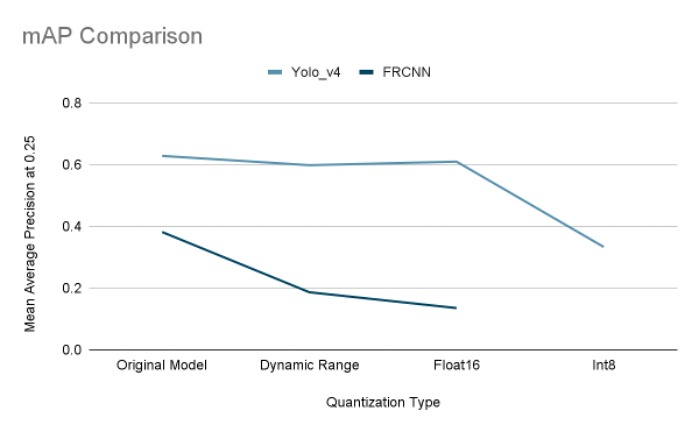}
  \caption{mAP@0.25 of Object Detection Model}
  \label{fig:img6}
\end{figure}

\begin{figure}[!t]
\centering 
  \includegraphics[width=2.5in]{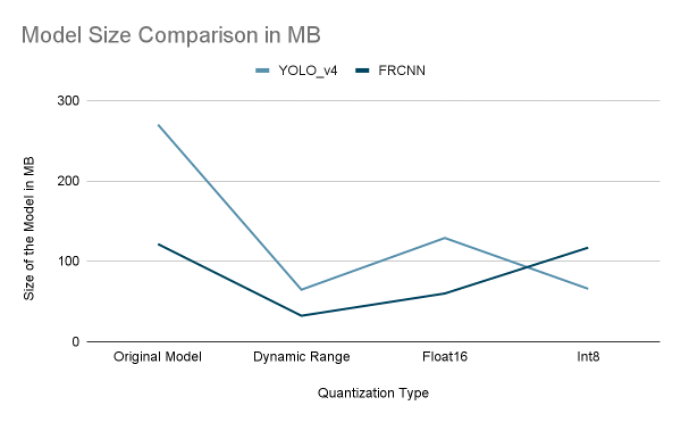}
  \caption{Model Size Comparison of Object Detection Model}
  \label{fig:img7}
\end{figure}

\begin{table}[!t]
\caption{Quantization Results of MobileNet Model  \label{tab:table3}}
\centering

\scalebox{0.5}{\begin{tabular}{|c||c||c||c||c||c||c|} 

 \hline
   & Size & Time to compress   & Time to load in sec. & Accuracy   &  Latency in sec.    &  FPS  \\
 \hline
 
 Original Model & 14.63 MB  & NA& ~ 0.71386   &  82.85  & ~0.0678  &  ~14.7563 \\
 \hline 
 
 Dynamic Range Quantized   &   3.92 MB  &  30 to 60   &       ~ 0.05826 &   71.52 &   ~0.2056 &  ~4.2528 \\
 \hline 
 
Float16 Quantized      &     7.03 MB    &    30 to 60   &      ~0.01241   &     73.2 &      ~ 0.0191  &       ~ 53.2336   \\

\hline 

Int8 Quantized    &     3.99 MB     &    120 to 145   &      0.001261   &      13.2484      &       ~1.01142  &          ~0.9887   \\
\hline
\end{tabular}}
\end{table}

\begin{figure}[!t]
\centering 
  \includegraphics[width=2.5in]{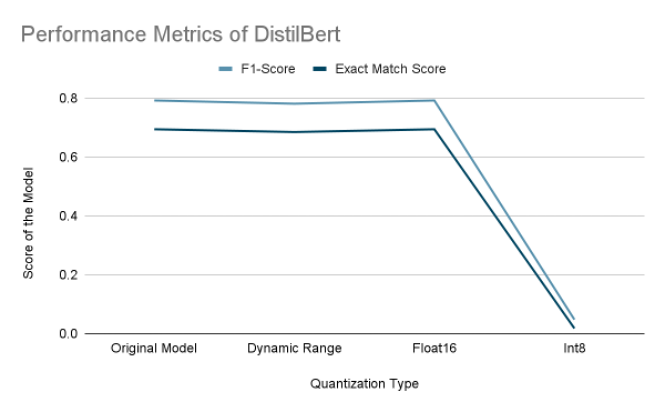}
  \caption{Performance Metrics Comparison of DistilBert Base Uncased Model}
  \label{fig:img8}
\end{figure}

\section{Results of Model Quantization}

\subsection{Image Classification Task}
We evaluated the pre-trained model using Imagenet’17 validation dataset \cite{20} with accuracy as the performance metrics. While the performance of all the three models differed by a small margin for dynamic range and float16 quantized model, it dropped by approximately 15\% for ResNet50 and $>75\%$ for MobileNet and Inception when optimized with int8 quantization Fig. \ref{fig:img4}. 
    Quantization results of MobileNet and Inception models are discussed in Table \ref{tab:table3} and Table\ref{tab:table4}. Also, as claimed by the official TensorflowLite documentation \cite{4}, the size of the models were reduced by almost 4x for dynamic range and int8 quantized models and halved for float16 quantized models Fig. \ref{fig:img5}.

\begin{table}[!t]
\caption{ Qunatization Results of Inception Model\label{tab:table4}}
\centering

\scalebox{0.5}{\begin{tabular}{|c||c||c||c||c||c||c|} 
  
 \hline
   & Size & Time to compress   & Time to load in sec. & Accuracy   &  Latency in sec.    &  FPS  \\
 \hline
 
 Original Model & 96.3 MB  & NA& ~ 1.1357  &  87.4394   & ~0.18885 &  ~5.295 \\
 \hline 
 
 Dynamic Range Quantized   &   24.3 MB  &  120 to 160    &       ~ 0.00072  &   87.51592  &   ~28.1448 &  ~0.03553 \\
 \hline 
 
Float16 Quantized      &     47.7 MB    &   130 to 170  &      ~0.001   &     84.4395 &      ~ 0.28421  &       ~ 4.029  \\

\hline 

Int8 Quantized    &     24.3 MB     &    120 to 160   &      ~0.01399   &      21.2993     &       ~39.74727   &          ~0.02515    \\
\hline
\end{tabular}}
\end{table}

\begin{figure}[!t]
\centering 
  \includegraphics[width=2.5in]{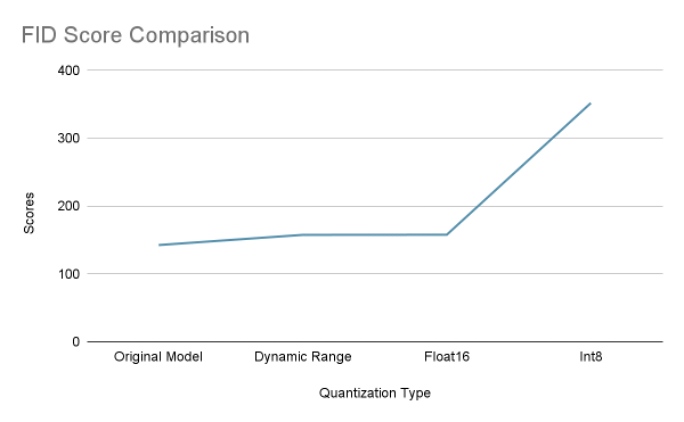}
  \caption{FID Score Comparison of CycleGAN Model}
  \label{fig:img9}
\end{figure}

\subsection{Object Detection Task}

\begin{table}[!t]
\caption{Quantization Results of Yolo\_v4 Model  \label{tab:table5}}
\centering

\scalebox{0.5}{\begin{tabular}{|c||c||c||c||c||c||c|} 

 \hline
    & Size & Time to compress in sec.   & Time to load in sec. & mAp0.25    &  Latency  in sec.  &  FPS  \\
 \hline
 
 Original Model & 269.9 MB  & NA& ~ 13.6985  &  ~0.628  & 0.55443  &  ~1.8036 \\
 \hline 
 
 Dynamic Range Quantized   &   64.7 MB  &     36.3951  &       ~ 0.00838   &   ~0.598 &   ~2.3040 &  ~0.43402  \\
 \hline 
 
Float16 Quantized      &     129 MB    &          34.919 sec.   &      ~0.05237   &     ~0.6094  &      ~ 1.11712  &       ~ 0.89515  \\

\hline 

Int8 Quantized    &     65.8 MB     &        78.4579 sec.   &      ~0.00796   &      ~0.3331      &       252.0263   &      ~0.00397   \\
\hline
\end{tabular}}
\end{table}

We tested the pre-trained object detection model on COCO’17 validation dataset \cite{24} with mean average precision at 0.25 as evaluation metric. While working with the FRCNN model we were able to optimize using dynamic range and float16 quantization, however, the results with the int8 optimization were not as expected. The model should have been reduced to one fourth of the original model but instead was only reduced by 4 MB Fig. \ref{fig:img7}. This also affected the model’s inference with the int8 quantized version as it gave a RunTimeError, hence we could not get the mAP for the same. Also due to the large inference time for some of the models, we only tested it on a subset of images from 3950 in the validation dataset. Apart from these exceptions, there was a minimal drop in the dynamic range and float16 quantized versions of YOLO\_v4 whereas the mAP dropped by $>45\%$ for int8 quantized versions Fig. \ref{fig:img6}. Quantization results of Yolo\_v4 and FRCNN are discussed in Table \ref{tab:table5} and Table \ref{tab:table6}.
\begin{table}[!t]
\caption{Quantization Results of FRCNN Model  \label{tab:table6}}
\centering

\scalebox{0.5}{\begin{tabular}{|c||c||c||c||c||c||c|} 
  
 \hline
   & Size & Time to compress   & Time to load & mAp0.25    &  Latency    &  FPS  \\
 \hline
 
 Original Model & 121.4 MB  & NA& ~ 7.16963 sec.  &  ~0.3815   & 1.30264 sec.  &  ~0.76767 \\
 \hline 
 
 Dynamic Range Quantized   &   32.3 MB  &     36.3951 sec.  &       ~ 0.00578 sec.   &   ~0.1865  &   ~250 sec. &  ~0.004  \\
 \hline 
 
Float16 Quantized      &     60 MB    &          34.919 sec.   &      ~0.03507 sec.   &     ~0.1353  &      ~ 4.002 sec.  &       ~ 0.02498   \\

\hline 

Int8 Quantized    &     117 MB     &        78.4579 sec.   &      ~0.02958 sec.   &      -        &       -     &           -     \\
\hline

\end{tabular}}
\end{table}

\subsection{Language Model}
We evaluated a pre-trained DistilBert Base Uncased Q\&A model \cite{26} taken from the hugging face library with the F-1 score and Exact Match score on the Squad v1.1 val dataset \cite{27}. The dynamic range and float16 quantized model size were ~x4 and ~x2 smaller respectively, than the original Table \ref{tab:table7} . Since the int8 optimized model took more than 42 seconds to process one query we only tested the model on 200 data points out of 20,964, hence, the result for the same remains inconclusive. Apart from these exceptions, the dynamic range quantized model showed no significant change in the performance whereas the float16 model did not change as compared to the original Fig. \ref{fig:img8}. Quantization results of DistilBert model are discussed in Table \ref{tab:table7}.

\begin{table}[!t]
\caption{Qunatization Results of DistilBert Model  \label{tab:table7}}
\centering
\scalebox{0.38}{\begin{tabular}{|c||c||c||c||c||c||c|}

\hline
   & Size & Time to compress in sec.   & Time to load in sec. & Average Prediction time in sec.   &  F1 Score    &  Exact Match  \\
 \hline
 
 Original Model & 266.7 MB  & NA& ~ 0.76894   &  ~0.33762   & 0.7917  &  0.69395 \\
 \hline 
 
 Dynamic Range Quantized   &   67.5 MB  &  7.86049    &       ~ 0.00265  &   ~1.2351 &   0.7808 &  0.68457  \\
 \hline 
 
Float16 Quantized      &     133 MB    &    7.60   &      ~0.00275   &     ~2.3224 &      0.7917  &       0.69395   \\

\hline 

Int8 Quantized    &     66.5 MB     &    327.337   &      ~0.002   &      ~42.39      &       0.04775 (200 data points)  &          0.01785    \\
\hline
\end{tabular}}
\end{table}

\subsection{Generative Model}

\begin{table}[!t]
\caption{ Quantization Results of CycleGAN Model  \label{tab:table8}}
\centering
\scalebox{0.45}{\begin{tabular}{|c||c||c||c||c||c||c|} 
 
 \hline
    & Size & Time to compress  in sec. & Time to load in sec. & Average Prediction Time in sec.   &  Average FID Score    \\
 \hline
 
 Original Model & 45.7 MB  & NA& ~ 0.3345  &  ~3.02771  &  142.569 \\
 \hline 
 
 Dynamic Range Quantized   &   11.6 MB  &     ~1.8787  &       ~ 0.00128   &   ~101.8902  & 157.558    \\
 \hline 
 
Float16 Quantized      &     22.7 MB    &    ~1.1981   &      ~0.00096   &     ~2.3451  &    157.759  \\

\hline 

Int8 Quantized    &     11.6 MB     &     ~85.6317  &      ~0.00173  &      ~1573.632    &  351.896\\
\hline
\end{tabular}}
\end{table}
We optimized and evaluated an open-source implementation of CycleGAN on horse-zebra dataset with the Fréchet inception distance (FID) score. The Fréchet inception distance (FID) is a metric used to assess the quality of images created by a generative model. Unlike the earlier inception score (IS), which evaluates only the distribution of generated images, the FID compares the distribution of generated images with the distribution of real images that were used to train the generator. The FID score between two similar images would be zero. Our model showed minimal change to the FID score for dynamic range and float16 quantized model whereas with the int8 optimized models the score worsened by more than 100\%, as compared to the original model Fig. \ref{fig:img9}. Quantization results of CycleGan model are discussed in Table \ref{tab:table8}
\def\thefootnote{*}\footnotetext{ MiniImageNet data is a sampled version of ImageNet data with 10 classes.}
\section{Results of Model Pruning}
\subsection{Image Classification}

In our experiments we have pruned ResNet50 and MobileNetv2 models pretrained on ImageNet dataset using tensorflow pruning toolkit \cite{17} and tested the pruned model on Mini ImageNet data$^*$. We have pruned these models in multiple stages and evaluated the change in performance as well as size of the model. The key idea of unstructured pruning is to utilize the fact that in a trained deep neural network majority of the weights have magnitudes around 0 and if we can prune these weights and finetune the model we can achieve somewhat similar performance to the original model. Fig. \ref{fig:img10} and Fig. \ref{fig:img11} represent distribution of weights in a particular layer of ResNet50 before and after pruning respectively.

Table \ref{tab:table9}  and Table \ref{tab:table10} show the results of pruning MobileNetV2 and ResNet50 models. Both the models seem to gain good performance as the networks are pruned. This could be due to the retraining of the network using the same data. However, for MobileNetv2 we see a drop in performance as we push the network to have a sparsity of around 90\%. Thus, it shows that pruning a model beyond a certain threshold can lead to an irreversible drop in performance.

\begin{table}[!t]
\caption{Pruning Results of MobileNetv2 \label{tab:table9}}
\centering
\scalebox{0.56}{\begin{tabular}{|c||c||c||c|} 
 \hline
     Pruning Percentage & Performance of Pruned Network & Size of Pruned Network & Inference Time of Pruned Network \\  
 \hline
 
 0\% & 85.27\%  & 27 Mb & 48.30ms. \\
 \hline 

  60\% & 88.10\%  & 9.2 Mb & 23.5ms.   \\
 \hline 
 
75\%    &    89.88\%    &    9.2 Mb   &      34.7ms.  \\

\hline 

90\%   &     84.88\%     &     9.2 Mb  &      35.12ms. \\
\hline
\end{tabular}}
\end{table}

\begin{table}[!t]
\caption{Pruning Results of ResNet50 \label{tab:table10}}
\centering
\scalebox{0.56}{\begin{tabular}{|c||c||c||c|} 

 \hline
     Pruning Percentage & Performance of Pruned Network & Size of Pruned Network & Inference Time of Pruned Network \\  
 \hline
 
 0\% & 78.85\%  & 271 Mb & 37.26ms. \\
 \hline 
 
  60 \% & 83.46\%  & 91 Mb & 18.56ms.   \\
 \hline 
 
75\%    &    87.00\%    &    91 Mb   &      36.63 ms.  \\

\hline 

90\%   &     86.06\%     &     91 Mb  &      37.53 ms. \\
\hline
\end{tabular}}
\end{table}

\subsection{Object Detection}

Similar to the image classification, we used tensorflow toolkit to prune the Yolo v4 object detection model pre-trained on COCO’17 dataset. The author of the original yolov4 paper had used different training and data augmentation techniques, however while retraining the model to achieve the desired sparsity we ran the whole dataset in coco’17 train folder for two epochs without following any data augmentation or any warm-up steps. In the results, we saw a sharp drop in the performance of the model and a slight increase in the inference time. The model size remained the same even after introducing sparsity as the unstructured pruning does not specifically remove the neuron or make any changes to the structure of the network.

\begin{table}[!t]
\caption{Pruning Results of Object Detection Model \label{tab:table11}}
\centering
\scalebox{0.6}{\begin{tabular}{|c||c||c||c||c|} 

 \hline
     Pruning Percentage & MAP@0.25 & Size  & Inference Time & FPS \\  
 \hline
 
 0\% & 0.628   & 259 Mb & 554 ms. & 1.8 \\
 \hline 
 
  60\% & 0.15  & 259 Mb & 638 ms. & 1.56  \\
 \hline 
 
80\%    &   0.14   &    259 Mb   &     623 ms.  & 1.6 \\
\hline
\end{tabular}}
\end{table}

\begin{figure}[!t]
\centering 
  \includegraphics[width=3.5in]{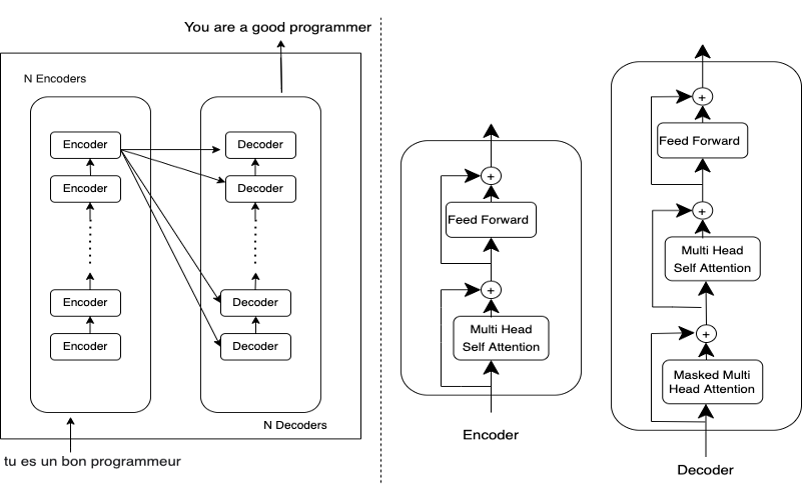}
  \caption{Transformer Architecture }
  \label{fig:img12}
\end{figure}

\begin{figure}[!t]
\centering 
  \includegraphics[width=2.5in]{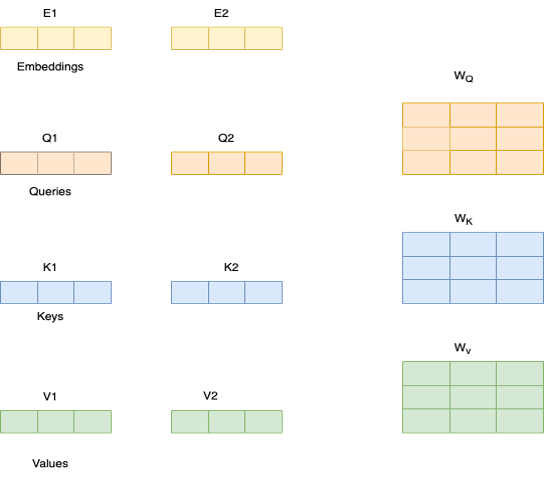}
  \caption{Inside the Attention Head }
  \label{fig:img13}
\end{figure}

\subsection{Language Models}

We have worked on XLM Roberta \cite{31} (XLM-R) and BERT-small  to analyze effects of structured pruning. Generally, transformers contain a stack of encoders followed by a stack of decoders. These encoders and decoders contain multi-head attention modules as well as feed forward networks. From a compression point of view, we can reduce the number of attention heads as well as optimize feed forward network using structured pruning. Fig. \ref{fig:img12} represents the detailed architecture of Transformer network and Fig. \ref{fig:img13} represents the building blocks of attention heads.

\begin{figure}[!t]
\centering 
  \includegraphics[width=2 in]{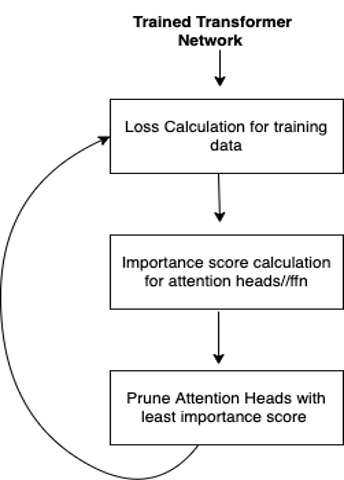}
  \caption{Pipeline for Transformer Pruning }
  \label{fig:img14}
\end{figure}

Most of the studies on transformers have shown that not all attention heads are equally important in the transformers, and some of the attention heads can be pruned without performance loss. Thus, Identifying and removing the least important attention heads can reduce the model size and have a small impact on performance. Along with the attention heads we also prune the feed forward network to achieve better compression results. So, the criterion for pruning is to calculate importance score of attention heads and feed forward networks neuron with respect to loss function calculated on the training data and remove those entities having lower importance score. Fig. \ref{fig:img14} presents the pipeline that has been followed to prune the transformer network.

For XLM-Roberta, PAWS-X, Cross lingual dataset \cite{32} has been used in sequence classification setting. Results of XLM-Roberta pruning have been discussed in Table \ref{tab:table12}. (N, H) in Network Architecture represents the number of neurons in hidden layers of feed forward unit and number of attention heads in the architecture. Original XLM-R network has 3072 neurons in feed forward network and 12 attention heads. Pruned network with N = 2048 and H = 8 has almost similar accuracy to the original network but improved inference time than the original one. Further pruning of the network impacts the accuracy significantly. Bert-small model has been trained on disaster tweets dataset\cite{50} and has also been pruned in structured manner. Original network has 512 neurons in FFN and 4 attention heads. Results of BERT-small pruning have been discussed in the Table \ref{tab:table13}.

We also worked on compression of GPT-2 small \cite{49} finetuned on E2E NLG \cite{novikova2017e2e} challenge and used parameter aware sparse training method proposed by Li \textit{et al.} \cite{li-etal-2022-pst} to prune the GPT-2 small model which uses both data free as well as well as data driven weight importance score calculation approaches to induce sparsity in the network. We evaluate performance of GPT-2 small on E2E NLG challenge by estimating 5 metrics i.e. BLEU, NIST, METEOR, $ROUGE_L$ and $CIDE_r$.
 We tested the network performance on various stages of sparsity and observed that even after inducing 90\% sparsity into the network its performance is comparable to the original fine tuned network on E2E NLG challenge. Compression results of GPT-2 small network have been discussed in Table \ref{tab:table14}.

We further explored lower rank adaptation based method proposed by Hu \textit{et al.} \cite{hu2021lora} which works by freezing the original model weights and injects trainable lower rank matrices in each layer of transformer network thus reducing the trainable parameters drastically. This idea can be hugely beneficial in multitasking where only one instance of original model needs to be deployed and multiple instances of task specific trainable lower rank modules can be deployed which will reduce the memory footprint drastically. We explored this idea on GPT-2 small finetuned on E2E NLG challenge and results are discussed in Table \ref{tab:table15}. Results suggest that while using lower rank i.e. 4 or 64 , the low rank module size is too small in comparison to original model  and we can deploy multiple such modules trained on various tasks and require only one instance of original network instead of multiple instances of original network finetuned on various tasks. 

Along with GPT-2 small we also evaluated three open-source large language models (LLMs) - Falcon-7B, LLaMA-2 7B, and Flan-T5-Large - for the task of summarization using the huggingface dataset "knkarthick/dialogsum". We compared the performance of the original models against their 4Bit quantized and LoRA tuned counterparts using Rouge, Bleu, and Meteor metrics and results are discussed in Table \ref{tab:table16}, Table \ref{tab:table17}, Table \ref{tab:table18}. Our findings indicate that the 4Bit quantized models achieved comparable performance to the original models across all evaluation metrics. This suggests that 4Bit quantization can significantly reduce model size and computational requirements without significantly sacrificing summarization quality. The LoRA optimization technique further improved the summarization performance of the LLM models, yielding higher Rouge, Bleu, and Meteor scores compared to the original and 4Bit quantized models. 

The implications of these findings are significant for LLM development in summarization tasks. The ability to optimize LLM models through techniques like 4Bit quantization and LoRA can lead to more efficient and scalable solutions. This enables faster fine-tuning times, reduced memory footprint, and improved resource utilization, making open-source LLM models more accessible and practical for real-world applications. Additionally, these optimization techniques pave the way for deploying LLMs on resource-constrained devices and systems, expanding their potential use cases.

\begin{table}[!t]
\caption{Pruning Results of XLM-Roberta \label{tab:table12}}
\centering
\scalebox{0.6}{\begin{tabular}{|c||c||c||c|} 

\hline
    Network Architecture (N,H) & Accuracy & Size of N/W & Inference Time (per batch) \\  
 \hline
 
 (3072,12) & 94.65\%  & 1.1 Gb& 2.106 sec. \\
 \hline 
 
  (2048,8) & 94.2\%  & 953 Mb & 1.385 sec.   \\
 \hline 
 
(1024,6)    &    85.15\%    &   863 Mb   &      0.788 sec.  \\

\hline 

(512,12)   &     48.05\%     &    881 Mb  &      1.04 sec. \\
\hline
\end{tabular}}
\end{table}

\begin{table}[!t]
\caption{Pruning Results of Bert-small \label{tab:table13}}
\centering
\scalebox{0.6}{\begin{tabular}{|c||c||c||c|} 

\hline
    Network Architecture (N,H) & Accuracy & Size of N/W & Inference Time (per batch) \\  
 \hline
 
 (512,4) & 95.80\%  & 19.2 Mb & 69.4 msec. \\
 \hline 
 
  (128,2) & 91.63\%  & 17.6 Mb & 40.94 msec.   \\
 \hline 
 
 (64,2)    &    90.69\%    &   17.3 Mb   &    44.18  msec.  \\
\hline 

 (256,1)    &    84.64\%    &   17.6 Mb   &    48.60  msec.  \\
\hline 

 (64,1)   &     79.79\%     &    16.8 Mb  &     25.88 msec. \\
\hline
\end{tabular}}
\end{table}

\begin{table}[!t]
\caption{Compression Results of GPT-2 Small \label{tab:table14}}
\centering
\scalebox{0.6}{
\begin{tabular}{|c||c||c||c||c||c|} 
\hline
    Sparsity & BlEU  & NIST & METEOR & ROUGE\_L & CIDEr  \\  
\hline
  $0\%$ & $0.6933$ & $8.8131$ & $0.4590$ & $0.7108$ &  $2.4643$   \\
\hline 
  $60\%$ & $0.7009$ & $8.8360$ & $0.4654$  & $0.7177$ &  $2.5170$  \\
\hline 
 $90\%$ & $0.6936$  & $8.7900$ & $0.4590$ & $0.7078$ & $2.4571$ \\
\hline 
\end{tabular}
}
\end{table}

\begin{table}[!t]
\caption{Low Rank Adaptation of GPT-2 Small on E2E Challenge \label{tab:table15}}
\centering
\scalebox{0.35}{
\begin{tabular}{|c||c||c||c||c||c||c||c||c||c|} 
\hline
Low Rank Dimension & Original No. of Parameters  & Trainable Parameters & Size of Original Model & Size of Low Rank Module & BLEU & NIST & METEOR & ROUGE\_L & CIDE\_r \\  
\hline
Full Finetuning & 124M & 124M & 523Mb & NA &  0.6682 & 8.58 & 0.4549  & 0.6765 & 2.28    \\
\hline 
r = 4  & 124M & 0.147M & 523Mb & 584Kb &  0.6489 & 8.4571 & 0.4345  & 0.6624 & 2.2412   \\
\hline 
r = 64  & 124M & 2.35M & 523Mb & 9.1Mb &  0.6656 & 8.5778 & 0.4438 & 0.6743 & 2.3027   \\
\hline 
\end{tabular}
}
\end{table}

\begin{table}[!t]
\caption{Compression Results of Flan-T5-Large \label{tab:table16}}
\centering
\scalebox{0.6}{\begin{tabular}{|c||c||c||c|} 

\hline 
Metric & Original & 4Bit & LoRA \\
\hline 
Average Inference time & 1.6145 & 2.9444 & 1.7672 \\
\hline
rouge1 & .2809 & .3317 & .3439 \\
\hline 
rouge2 & 0.0914 & 0.1222 & 0.1306 \\
\hline 
rougeL & 0.2364 & 0.2741 & 0.2885 \\
\hline 
rougeLsum & 0.2366 & 0.2742 & 0.2887 \\
\hline 
bleu & 0.0506 & 0.0982 & 0.0982 \\
\hline 
meteor & 0.1942 & 0.2777 & 0.2738 \\
\hline 

 \end{tabular}}
\end{table}

\begin{table}[!t]
\caption{Compression Results of Falcon-7B \label{tab:table17}}
\centering
\scalebox{0.6}{\begin{tabular}{|c||c||c||c|} 

\hline 
Metric & Original & 4Bit & LoRA \\
\hline 
Average Inference time & 4.7768 & 6.7933 & 6.4898 \\
\hline
rouge1 & .0911 & .0908 & .0.1696 \\
\hline 
rouge2 & 0.0241 & 0.0242 & 0.0395 \\
\hline 
rougeL & 0.0738 & 0.0744 & 0.1263 \\
\hline 
rougeLsum & 0.0797 & 0.0807 & 0.1446 \\
\hline 
bleu & 0.0137 & 0.0135 & 0.0216 \\
\hline 
meteor & 0.2133 & 0.2184 & 0.3155 \\
\hline 

 \end{tabular}}
\end{table}

\begin{table}[!t]
\caption{Compression Results of LLaMA-2-7B \label{tab:table18}}
\centering
\scalebox{0.6}{\begin{tabular}{|c||c||c||c|} 

\hline 
Metric & Original & 4Bit & LoRA \\
\hline 
Average Inference time & 7.5620 & 20.5844 & 7.8099 \\
\hline
rouge1 & .0926 & .0928 & .1696 \\
\hline 
rouge2 & 0.0212 & 0.0219 & 0.0395 \\
\hline 
rougeL & 0.0799 & 0.0798 & 0.1263 \\
\hline 
rougeLsum & 0.0768 & 0.0768 & 0.1446 \\
\hline 
bleu & 0.0113 & 0.0117 & 0.0216 \\
\hline 
meteor & 0.2217 & 0.2260 & 0.3151 \\
\hline 

 \end{tabular}}
\end{table}

\subsection{ Generative Model}
\begin{figure}[!t]
\centering 
  \includegraphics[width=2.5in]{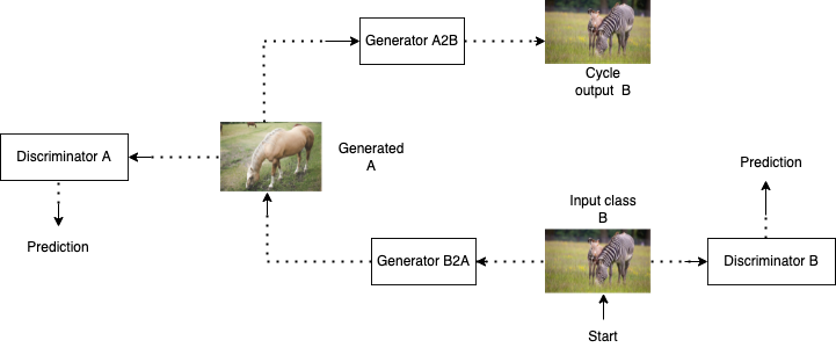}
  \caption{General flow of CycleGAN}
  \label{fig:img15}
\end{figure}

\begin{figure}[!t]
\centering 
  \includegraphics[width=2.5in]{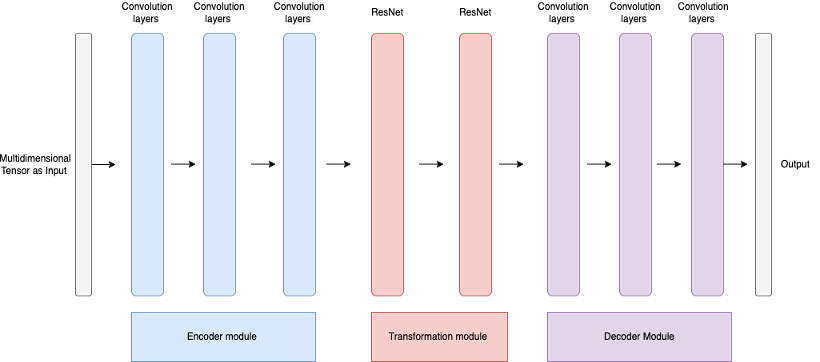}
  \caption{Generator Network}
  \label{fig:img16}
\end{figure}

For generative models we have used cyclegan for compression and evaluated the compression results using channel pruning and knowledge distillation as a single optimization task. The architecture of cyclegan consists of 2 generators as well as 2 discriminators for two tasks A2B and B2A conversion where A and B are 2 different domains as represented in Fig. \ref{fig:img15}. Once the cyclegan gets trained we only deploy generators so from a compression point of view our only interest lies in generators. Generator architecture consists of encoding convolutional layers followed by a stack of residual blocks then few decoding convolutional layers as shown in Fig. \ref{fig:img16}. 

Here we prune the channels of the original generator network along with distillation of knowledge from original to pruned one. So, the optimization function for this compression problem is given by (\ref{eq:1}) where \begin{math} L_{GAN} \end{math}  represents standard GAN loss. We minimize \begin{math} L_{GAN} \end{math}  with respect to generator G and maximize with respect to discriminator D.

\begin{figure}
     \centering
     \subfloat[][]{\includegraphics[width=.716in]{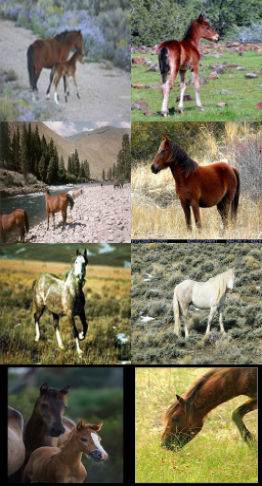}\label{}}
     \subfloat[][]{\includegraphics[width=.79in]{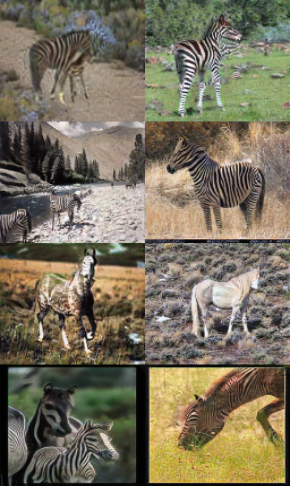}\label{}}
     \subfloat[][]{\includegraphics[width=.75in]{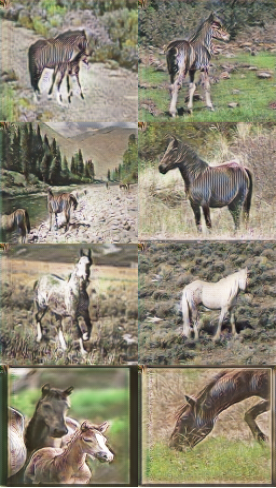}\label{}}
     \caption{Comparison of results for CycleGan Compression. (a) Original Images (b) Generated Images from Original CycleGan (c) Generated Images from Pruned CycleGan}
     \label{steady_state}
\end{figure}

\begin{figure}[!t]
\centering 
  \includegraphics[width=2.5in]{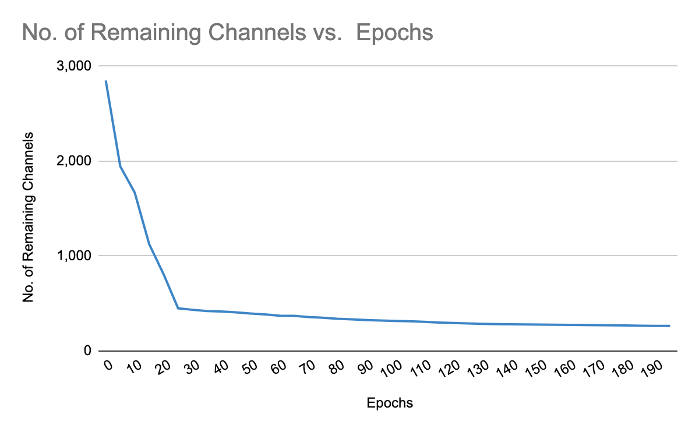}
  \caption{Number of Remaining Generator Channels vs. Epochs}
  \label{fig:img18}
\end{figure}

\begin{figure}[!t]
\centering 
  \includegraphics[width=2.5in]{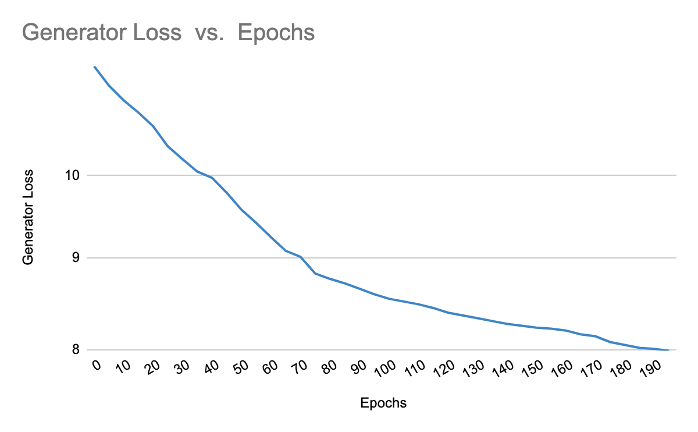}
  \caption{Total Generator loss vs. Epochs }
  \label{fig:img19}
\end{figure}

\begin {equation} \label{eq:1}  \min_{G} \max_{D}   L_{GAN} = E_{v \in V}[log(D(v)]  + E_{u \in U}[log(1-D(G(u)))]\end{equation}

Where D is discriminator jointly trained with efficient generator G where G converts images from domain U to domain V and u and v represents images from domain U and V respectively. \begin{math} E(x) \end{math}represents expected value of the random variable x.

\begin{equation}\label{eq:2} L_{distill} = E_{u \in U} [d(G(u),G_{0}(u))] \end{equation}

 Another distillation loss \begin{math} L_{distill} \end{math}  is used to enforce the compressed generator G to mimic G0, the original generator as given by (\ref{eq:2}) where \begin{math} d(x,y) \end{math} is some distance metric to compute distance between data points x and y.

 \begin{equation}\label{eq:(3)} L_{distill}(W,\phi) + \delta L_{c}(\phi) = E[d((G(u;W,\phi),G_{0}(u))] + \delta \lVert \phi \rVert_{1} \end{equation}

The idea is to get the architecture of G from \begin{math}G_{0}\end{math} through channel pruning which is done using L1 norm on the trainable scale parameter $\phi$ in the normalization layers as given in (\ref{eq:(3)}) where \begin{math} L_{c}(\phi)  \end{math} represents \begin{math} L_{1} \end{math} norm of $\phi$ and $\delta$ is the trade-off parameter controlling the network sparsity level. W represents trainable weight parameters in generator G.

\begin{equation} \label{eq:4} L(W,\phi,\alpha) = L_{GAN}(W,\phi,\alpha) + \epsilon L_{distill}(W,\phi) + \delta L_{c}(\phi) \end{equation}

Equation (\ref{eq:4}) represents the total GAN loss L where $\alpha$ represents parameters in discriminator D and $\epsilon$ is a hyperparameter which controls the weightage to the distillation process in the algorithm.

CycleGan model with 9 residual blocks has been compressed using channel pruning and simultaneous knowledge updation from the original trained network and results have been observed on horse - zebra dataset. Original model size was 44 Mb and it was compressed up to 7.7Mb while FID (Fréchet Inception Distance) score calculated for pruned network was 122.29 while for the original network it was 74.04. Inference time measured for the original network was 6.81 ms. for a batch of 8 images while inference time for pruned network was 5.8 ms. Fig. \ref{steady_state} represents a comparison between original cyclegan output vs. compressed cyclegan network whereas Fig. \ref{fig:img18} and Fig. \ref{fig:img19} represents no. of channels remaining in generator network after iterative pruning vs. epochs and total generator loss vs. number of epochs during compression phase and respectively.
Extending this experiment further, pruning with knowledge transfer and quantization(8bit) was also explored as a single joint optimization problem rather than performing quantization post training which generally results in poor performance. While performing all the three optimizations jointly, original model size was 44 Mb and it was compressed up to 4 Mb while FID score calculated for pruned network was 153.98 while for the original network it was 74.04. Inference time measured for the original network was 6.65ms. for a batch of 8 images while inference time for the pruned network was 5.59 ms. for the same batch size.

\section{observations}

With the experimental method for quantization and pruning mentioned in section [2.2], we observed that:
\subsection{Quantization}

\begin{description}

\item[$\bullet$] Upon optimizing a Tensorflow based DNN model using Dynamic range quantization, the model size is reduced by almost a quarter of its original size.

\item[$\bullet$] Upon optimizing a Tensorflow based DNN model using Float16 quantization, the model size is reduced to almost half of its original size.

\item[$\bullet$] The latency of quantized models is much higher than the original. Although the int8 quantized model should have the lowest latency as it performs operations on 8-bit integer values, this is not observed while inference on the intel’s i-9 processor. For some models, latency for the int8 quantized version was more than 450 times the latency of the original Table \ref{tab:table7}. This abnormality can be explained by the Instruction Set Extensions \cite{35} that each processor follows. Instruction Set Extensions are additional instructions which can increase performance when the same operations are performed on multiple data objects. Intel’s i9-988oH follows Intel® SSE4.1, Intel® SSE4.2, Intel® A VX2 extensions \cite{30} whereas TensorflowLite utilizes CPU kernels that are optimized for ARM Neon Architecture \cite{36}\cite{37}. TensorFlow Lite's Delegate API acts as a bridge between the TFLite runtime and the lower-level APIs to handle the heavy arithmetic that is typically found in Machine Learning models. Delegates \cite{37} enable hardware acceleration of TensorFlow Lite models by leveraging on-device accelerators such as the GPU and Digital Signal Processor. Hence, due to these incompatibilities, the quantized models have relatively higher latency than the original models on intel’s i9-988oh processor.

\item[$\bullet$] While quantizing pre-trained models, using different Tensorflow versions affected the size of compressed models by a few MBs. Also, quantized models from one version of Tensorflow might not be compatible with another version.

\item[$\bullet$] 
Despite the standardization of the quantization method from the Tensorflow, not all operations of a neural network can be quantized as TensorflowLite only follows a subset of operations by Tensorflow and with some limitations \cite{46}.

\item[$\bullet$] 
For some models, the input and output shape as well as sequence of the converted tflite model might not be the same as the original one and the expected parameters need to be provided at the time of quantization \cite{47}.

\end{description}

\subsection{Pruning}

\begin{description}
\item[$\bullet$]  A lot of overparameterized pre-trained deep learning models can be significantly pruned to increase the sparsity and reduce the number of parameters while having the least effect on the performance.
 
\item[$\bullet$] The pruned model when retrained with the original data may increase the performance as compared to the original Table \ref{tab:table9} and Table \ref{tab:table10}.

\item[$\bullet$] There is a threshold till which the models can be pruned efficiently without any significant effect on the performance Table \ref{tab:table12}. Pruning models beyond this threshold may lead to irreversible change in the performance i.e., the model would not have enough parameters to infer correctly from the data.

\item[$\bullet$]

Unstructured pruning of a deep learning model alone does not necessarily reduce the size of the model as the weights that are reduced to zero still occupy 32 bit of float value which can be either eliminated or quantized to lower the memory requirement for execution.

\item[$\bullet$]

Structured pruning removes entire neurons and the associated output connections and hence a reduction in size of the model is observed wherever this method is applied.

\item[$\bullet$]

Pruning does not guarantee any decrease in the latency, and neither is observed in our experiments. Both the frameworks i.e. TensorFlow \cite{17} and PyTorch have only mentioned about the reduction in model’s performance with respect to the increase in sparsity only. In one of the technical forums \cite{40} PyTorch developers does mention that the pruning features are experimental and do not necessarily guarantee inference time speed up or reduced memory requirements and similar response is seen for Tensorflow \cite{38}.

\item[$\bullet$]
The Tensorflow currently supports only the functional and sequential keras models \cite{39}; a broader set of models is anticipated to be supported in upcoming versions of Tensorflow i.e. TF $> 2.9 a$.

\end{description}

\subsection{Recommendation}

\begin{description} 
\item[$\bullet$]
Quantization method is best suited for edge deployment use cases where the microprocessors or microcontrollers are optimized to perform integer-based calculations.

\item[$\bullet$]

For pre-trained deep learning models, we can expect the size to reduce to half and to its quarter while quantizing it to float16 and int8, respectively, using Tensorflow model optimization toolkit.

\item[$\bullet$]

Resnet, MobileNet and VGG16 based networks can be quantized or pruned effectively using Tensorflow.

\item[$\bullet$]
A little drop in performance can be expected if quantized the model, however, pruning may drastically reduce the performance of a network.

\item[$\bullet$]

Pruning is a subjective process, and its efficiency is directly related to the number of insignificant weights in the model. Suppose we have a neural network where most of the neurons are apt in extracting features from the data and contribute to the final prediction. If we try to prune this model by even a small amount, then a large drop in performance can be expected. Similarly, if we have a bigger network for any simple task in which the majority of neurons are hardly contributing to the output then this model can be pruned significantly while having a minimum performance change. This was witnessed while pruning the RoBerta model.

\item[$\bullet$]

Quantizing a model post pruning is recommended for the edge use cases.

\item[$\bullet$]
Not all the models can be pruned or quantized as the custom layers and connections implemented by them might not yet be supported by the tensorflow framework.

\end{description}

\section{Conclusion}

The tensorflow based models can be optimized using post training quantizations and the optimized models have reduced model size. The performance of a quantized model is affected by the size of the original model, bigger the original model the quantization of parameters will have lesser effect on the performance as compared to a shallow network which already has a relatively lower number of parameters to predict from. The different quantized versions of a model might have a better inference speed on processors that are optimized for integer-based operations but it has significantly higher inference time on intel’s i9 processor. As by default, TensrflowLite utilizes CPU kernels that follows ARM Neon Architecture, the TFLite quantized model will work best with the devices like Google Coral \cite{41}, Raspberry Pi (2, 3B, 4B) [42], Jetson Nano \cite{43} and others which have ARM Cortex-A \cite{44}and ARM Cortex-R \cite{45} family processors.

Pruning insignificant weights is an effective way of reducing the model to the bare minimum requirement of producing the desired results. This is an active area of research and in our experiments, we used the approaches provided by the Tensorflow toolkit as well as from some of the recent papers. As of now, the Tensorflow toolkit does not support the pruning of non Keras networks and has trouble handling bigger networks. They do not claim any reduction in the size of the model in their official documentation but only present the change in inference time of the models when they are quantized after pruning. In our experiments too we witnessed somewhat similar results. Aggressive pruning to increase the sparsity of a model might result in significant drop in performance which is not desirable and hence it is recommended to gradually increase the pruning parameters. For the edge deployment use cases, we can further quantize this pruned network to reduce the size and make them compatible with the edge hardware architecture. It is also recommended to implement pruning as a part of the training process so that the network size shrinks as the model learns and keeps only the required neurons in the final version. This would save a lot of time as the pruning after training takes the equal amount of time as the network has to go through all the training data again.

\bibliography{paper_fv}

\begin{thebibliography}{10}
\providecommand{\url}[1]{#1}
\csname url@samestyle\endcsname
\providecommand{\newblock}{\relax}
\providecommand{\bibinfo}[2]{#2}
\providecommand{\BIBentrySTDinterwordspacing}{\spaceskip=0pt\relax}
\providecommand{\BIBentryALTinterwordstretchfactor}{4}
\providecommand{\BIBentryALTinterwordspacing}{\spaceskip=\fontdimen2\font plus
\BIBentryALTinterwordstretchfactor\fontdimen3\font minus
  \fontdimen4\font\relax}
\providecommand{\BIBforeignlanguage}[2]{{%
\expandafter\ifx\csname l@#1\endcsname\relax
\typeout{** WARNING: IEEEtran.bst: No hyphenation pattern has been}%
\typeout{** loaded for the language `#1'. Using the pattern for}%
\typeout{** the default language instead.}%
\else
\language=\csname l@#1\endcsname
\fi
#2}}
\providecommand{\BIBdecl}{\relax}
\BIBdecl

\bibitem{1}
J.~Dean, ``{Google Research: Themes from 2021 and Beyond},''
  \url{https://ai.googleblog.com/2022/01/google-research-themes-from-2021-and.html},
  [Google AI Blog, 11 January 2022].

\bibitem{2}
C.~Ha, ``{Convergence of SoTA CV models. The year is 2021 and the
  recent{\ldots} | by Chris Ha},''
  \url{https://medium.com/@hac541309/convergence-of-sota-cv-models-ad985a597173},
  [Medium, 26 September 2021].

\bibitem{3}
``{IEEE 754},'' \url{https://en.wikipedia.org/wiki/IEEE_754}, [Wikipedia,
  2000].

\bibitem{4}
Tensorflow, ``{Post-training quantization},''
  \url{https://www.tensorflow.org/lite/performance/post_training_quantization},
  [TensorFlow, 26 May 2022].

\bibitem{5}
Pytorch, ``{Quantization --- PyTorch 1.12 documentation},''
  \url{https://pytorch.org/docs/stable/quantization.html}, [PyTorch, 2019].

\bibitem{6}
Github, ``{caffe/quantization.cpp at master · intel/caffe · GitHub},''
  \url{https://github.com/intel/caffe/blob/master/src/caffe/quant/quantization.cpp},
  [GitHub, 2018].

\bibitem{7}
B.~Jacob, S.~Kligys, B.~Chen, M.~Zhu, M.~Tang, A.~Howard, H.~Adam, and
  D.~Kalenichenko, ``Quantization and training of neural networks for efficient
  integer-arithmetic-only inference,'' in \emph{Proceedings of the IEEE
  conference on computer vision and pattern recognition}, 2018, pp. 2704--2713.

\bibitem{8}
R.~Krishnamoorthi, ``{[1806.08342] Quantizing deep convolutional networks for
  efficient inference: A whitepaper},'' \url{https://arxiv.org/abs/1806.08342},
  [arXiv, 21 June 2018].

\bibitem{9}
M.~M. T.~Dubhir and R.~Singhal, ``{Benchmarking of Quantization Libraries in
  Popular Frameworks},'' [2021 IEEE International Conference on Big Data (Big
  Data), 2021, pp. 3050-3055].

\bibitem{10}
C.~Luo, ``{[2005.05085] Comparison and Benchmarking of AI Models and Frameworks
  on Mobile Devices},'' \url{https://doi.org/10.48550/arXiv.2005.05085},
  [arXiv, 7 May 2020].

\bibitem{11}
Tensorflow, ``{Performance measurement},''
  \url{https://www.tensorflow.org/lite/performance/measurement}, [TensorFlow,
  16 April 2022].

\bibitem{12}
D.~Blalock, ``{[2003.03033] What is the State of Neural Network Pruning?}''
  \url{https://doi.org/10.48550/arXiv.2003.03033}, [arXiv, 6 March 2020].

\bibitem{13}
S.~Srinivas and R.~Venkatesh, ``{[1507.06149] Data-free parameter pruning for
  Deep Neural Networks},'' \url{https://arxiv.org/abs/1507.06149}, [arXiv, 22
  July 2015].

\bibitem{14}
Tensorflow, ``{FlatBuffers: FlatBuffers},''
  \url{https://google.github.io/flatbuffers/}, [Google, 2020].

\bibitem{15}
Wikipedia, ``{Half-precision floating-point format},''
  \url{https://en.wikipedia.org/wiki/Half-precision_floating-point_format},
  [Wikipedia, 2000].

\bibitem{16}
V.~Lendave, ``{A Beginner's Guide to Neural Network Pruning},''
  \url{https://analyticsindiamag.com/a-beginners-guide-to-neural-network-pruning/},
  [Analytics India Magazine, 18 September 2021].

\bibitem{17}
Tensorflow, ``{Trim insignificant weights},''
  \url{https://www.tensorflow.org/model_optimization/guide/pruning},
  [TensorFlow, 3 August 2022].

\bibitem{18}
Z.~Yang, ``{[2203.15996] TextPruner: A Model Pruning Toolkit for Pre-Trained
  Language Models},'' \url{https://arxiv.org/abs/2203.15996}, [arXiv, 30 March
  2022].

\bibitem{19}
H.~Wang, ``{[2008.11062] GAN Slimming: All-in-One GAN Compression by A Unified
  Optimization Framework},'' \url{https://arxiv.org/abs/2008.11062}, [arXiv, 25
  August 2020].

\bibitem{20}
Imagenet, ``{ImageNet Large Scale Visual Recognition Challenge 2017
  (ILSVRC2017},'' \url{https://image-net.org/challenges/LSVRC/2017/},
  [ImageNet, 2017].

\bibitem{21}
Keras, ``{Keras Applications},'' \url{https://keras.io/api/applications/},
  [Keras, 2017].

\bibitem{22}
V.~Hung, ``{hunglc007/tensorflow-yolov4-tflite: YOLOv4, YOLOv4-tiny, YOLOv3,
  YOLOv3-tiny Implemented in Tensorflow 2.0, Android. Convert YOLO v4 .weights
  tensorflow, tensorrt and tflite},''
  \url{https://github.com/hunglc007/tensorflow-yolov4-tflite}, [GitHub, 2020].

\bibitem{23}
Tensorflow, ``{TensorFlow Hub},''
  \url{https://tfhub.dev/tensorflow/faster_rcnn/resnet50_v1_640x640/1},
  [TensorFlow Hub, 2018].

\bibitem{24}
T.-Y. Lin, ``{[1405.0312] Microsoft COCO: Common Objects in Context},''
  \url{https://arxiv.org/abs/1405.0312}, [arXiv, 1 May 2014].

\bibitem{25}
X.~Xu, ``{YOLO vs Faster RCNN},''
  \url{https://everitt257.github.io/post/2018/08/10/object_detection.html},
  [Everitt's blog, 10 August 2018].

\bibitem{26}
HuggingFace-PyTorch, ``{distilbert-base-uncased-distilled-squad· Hugging
  Face},''
  \url{https://huggingface.co/distilbert-base-uncased-distilled-squad},
  [Hugging Face, 1 July 2022].

\bibitem{27}
P.~Rajpurkar, ``{[1606.05250] SQuAD: 100,000+ Questions for Machine
  Comprehension of Text},'' \url{https://arxiv.org/abs/1606.05250}, [arXiv, 16
  June 2016].

\bibitem{28}
L.~Ho, ``{LynnHo/CycleGAN-Tensorflow-2},''
  \url{https://github.com/LynnHo/CycleGAN-Tensorflow-2}, [GitHub, 2020].

\bibitem{29}
U.~Berkeley, ``{Index of /\~taesung\_park/CycleGAN/datasets},''
  \url{https://people.eecs.berkeley.edu/\~taesung_park/CycleGAN/datasets/},
  [People @ EECS at UC Berkeley, 2017].

\bibitem{31}
A.~Conneau, ``{[1911.02116] Unsupervised Cross-lingual Representation Learning
  at Scale},'' \url{http://arxiv.org/abs/1911.02116}, [arXiv, 5 November 2019].

\bibitem{51}
Google, ``{Bert-Small},'' \url{https://huggingface.co/prajjwal1/bert-small}.

\bibitem{50}
Kaggle, ``{Disaster Tweet Dataset},''
  \url{https://www.kaggle.com/competitions/nlp-getting-started}.

\bibitem{49}
OpenAI, ``{GPT-2},'' \url{https://openai.com/blog/tags/gpt-2/}, 2019.

\bibitem{novikova2017e2e}
J.~Novikova, O.~Du{\v{s}}ek, and V.~Rieser, ``The {E2E} dataset: New challenges
  for end-to-end generation,'' in \emph{Proceedings of the 18th Annual Meeting
  of the Special Interest Group on Discourse and Dialogue}, 2017.

\bibitem{30}
Intel, ``{Intel Core i99880H Processor 16M Cache up to 4.80 GHz Product
  Specifications},''
  \url{https://ark.intel.com/content/www/us/en/ark/products/192987/intel-core-i99880h-processor-16m-cache-up-to-4-80-ghz.html},
  [Intel ARK, 2019].

\bibitem{48}
M.~Azure, ``{NCv3-series - Azure Virtual Machines.}''
  \url{https://docs.microsoft.com/en-us/azure/virtual-machines/ncv3-series},
  [Microsoft Docs, 24 May 2022].

\bibitem{32}
Y.~Yang, ``{[1908.11828] PAWS-X: A Cross-lingual Adversarial Dataset for
  Paraphrase Identification},'' \url{https://arxiv.org/abs/1908.11828}, [arXiv,
  30 August 2019].

\bibitem{li-etal-2022-pst}
Y.~Li, F.~Luo, C.~Tan, M.~Wang, S.~Huang, S.~Li, and J.~Bai,
  ``Parameter-efficient sparsity for large language models fine-tuning,'' in
  \emph{31th International Joint Conference on Artificial Intelligence}, 2022.

\bibitem{hu2021lora}
E.~J. Hu, Y.~Shen, P.~Wallis, Z.~Allen-Zhu, Y.~Li, S.~Wang, L.~Wang, and
  W.~Chen, ``Lora: Low-rank adaptation of large language models,'' in
  \emph{arXiv preprint arXiv:2106.09685}, 2021.

\bibitem{35}
Intel, ``{Intel{\textregistered} Instruction Set Extensions Technology},''
  \url{https://www.intel.com/content/www/us/en/support/articles/000005779/processors.html},
  [Intel, 2019].

\bibitem{36}
Neon-ARM, ``{Neon -- Arm{\textregistered}},''
  \url{https://www.arm.com/technologies/neon}, [Arm, 2020].

\bibitem{37}
Tensorflow, ``{TensorFlow Lite Delegates},''
  \url{https://www.tensorflow.org/lite/performance/delegates}, [TensorFlow, 30
  January 2021].

\bibitem{46}
``{TensorFlow Lite and TensorFlow operator compatibility},''
  \url{https://www.tensorflow.org/lite/guide/ops_compatibility}, [TensorFlow,
  28 January 2021].

\bibitem{47}
Stackoverflow, ``{Converting Mobilenet Model to TFLite changes input size},''
  \url{https://stackoverflow.com/questions/66404756/converting-mobilenet-model-to-tflite-changes-input-size},
  [Stack Overflow, 28 February 2021].

\bibitem{40}
Github, ``{Effectiveness of pruning · Issue 32928· pytorch/pytorch·
  GitHub},'' \url{https://github.com/pytorch/pytorch/issues/32928}, [GitHub, 3
  February 2020].

\bibitem{38}
``{Sparsity Runtime Integration with TF/TFLite for Latency Improvements· Issue
  173 · tensorflow/model-optimization},''
  \url{https://github.com/tensorflow/model-optimization/issues/173}, [GitHub, 6
  December 2019].

\bibitem{39}
``{Pruning: Keras subclassed model increased support · Issue 155 ·
  tensorflow/model-optimization},''
  \url{https://github.com/tensorflow/model-optimization/issues/155}, [GitHub,
  11 November 2019].

\bibitem{41}
Coral, ``{Products},'' \url{https://coral.ai/products/#prototyping-products},
  [Coral.ai, 2018].

\bibitem{43}
N.~Jetson, ``{Jetson Nano Developer Kit},''
  \url{https://developer.nvidia.com/embedded/jetson-nano-developer-kit},
  [NVIDIA Developer, 2018].

\bibitem{44}
Wikipedia, ``{ARM Cortex-A},''
  \url{https://en.wikipedia.org/wiki/ARM_Cortex-A}, [Wikipedia, 2018].

\bibitem{45}
``{ARM Cortex-M},'' \url{https://en.wikipedia.org/wiki/ARM_Cortex-M},
  [Wikipedia, 2018].

\end{thebibliography}
\bibliographystyle{IEEEtran}

\end{document}